\pdfoutput=1

\documentclass[11pt]{article}
\usepackage{authblk}
\usepackage[]{ACL2023}

\usepackage{times}
\usepackage{url}
\usepackage{latexsym}
\usepackage{amssymb}

\usepackage[T1]{fontenc}

\usepackage[utf8]{inputenc}

\usepackage{microtype}
\usepackage{amsmath}
\usepackage{multirow}
\usepackage{booktabs}
\usepackage{graphicx}
\usepackage{float}
\usepackage{hyperref}
\usepackage{algorithm}           
\usepackage{algorithmic}
\usepackage{tabularx}
\usepackage{listings}
\usepackage{array}

\title{S$^3$HQA: A Three-Stage Approach for Multi-hop Text-Table Hybrid Question Answering}

\author{
\textbf{Fangyu Lei}\textsuperscript{1,2}, 
\textbf{Xiang Li}\textsuperscript{1,2}, 
\textbf{Yifan Wei}\textsuperscript{1,2}, \\
\textbf{Shizhu He}\textsuperscript{1,2},
\textbf{Yiming Huang}\textsuperscript{1,2}, 
\textbf{Jun Zhao}\textsuperscript{1,2}, 
\textbf{Kang Liu}\textsuperscript{1,2} \\
  \textsuperscript{1}The Laboratory of Cognition and Decision Intelligence for Complex Systems,\\
  Institute of Automation, Chinese Academy of Sciences \\
  \textsuperscript{2}School of Artificial Intelligence, University of Chinese Academy of Sciences \\
\texttt{\{leifangyu2022, lixiang2022, weiyifan2021\}@ia.ac.cn}\\
\texttt{\{shizhu.he, jzhao, kliu\}@nlpr.ia.ac.cn}}

\begin{document}
\maketitle
\begin{abstract}
Answering multi-hop questions over hybrid factual knowledge from the given text and table (TextTableQA) is a challenging task. Existing models mainly adopt a retriever-reader framework, which have several deficiencies, such as noisy labeling in training retriever, insufficient utilization of heterogeneous information over text and table, and deficient ability for different reasoning operations. In this paper, we propose a three-stage TextTableQA framework \textbf{S$^3$HQA}, which comprises of \emph{retriever}, \emph{selector}, and \emph{reasoner}. We use a \emph{retriever with refinement training} to solve the noisy labeling problem. Then, a \emph{hybrid selector} considers the linked relationships between heterogeneous data to select the most relevant factual knowledge. For the final stage, instead of adapting a reading comprehension module like in previous methods, we employ a \emph{generation-based reasoner} to obtain answers. This includes two approaches: a row-wise generator and an LLM prompting generator~(first time used in this task). The experimental results demonstrate that our method achieves competitive results in the few-shot setting. When trained on the full dataset, our approach outperforms all baseline methods, ranking first on the HybridQA leaderboard.\footnote{\url{https://codalab.lisn.upsaclay.fr/competitions/7979}.} 

\end{abstract}

\section{Introduction}
\begin{figure}
    \centering
	\includegraphics[width=0.48\textwidth]{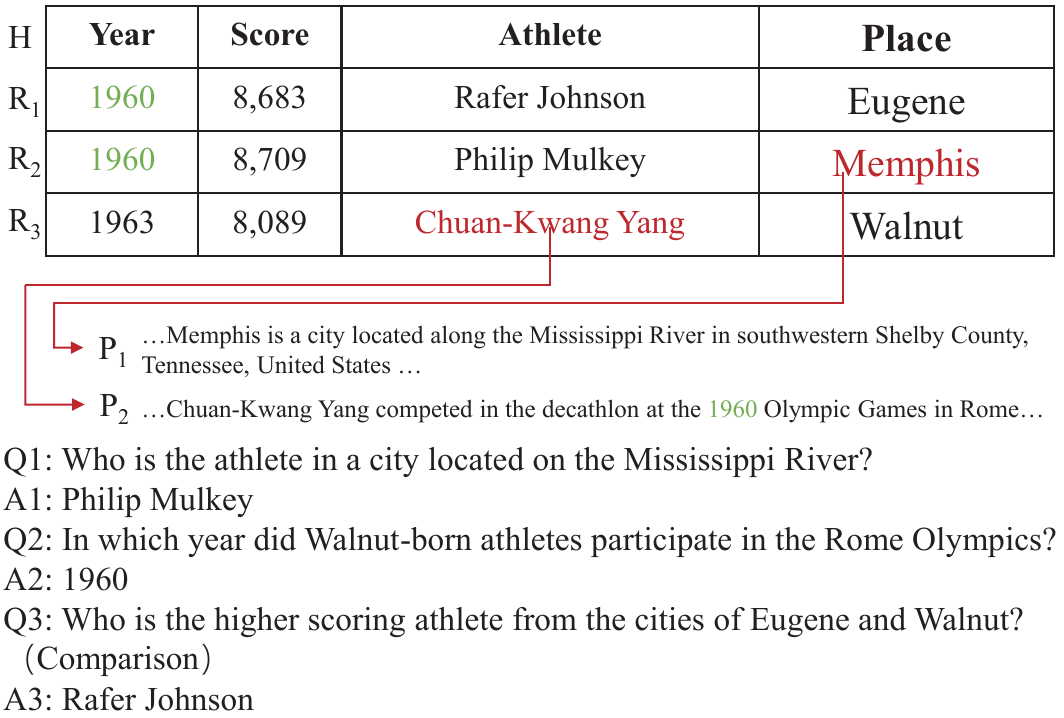}
	\caption{The examples of HybridQA.}
	\label{fig1}
\end{figure}

Question answering systems devote to answering various questions with the evidence located in the structured knowledge base (e.g., table) ~\citep{pasupat2015compositional, yu2018spider} or unstructured texts~\citep{rajpurkar2016squad}. Considering that many questions need to utilize multiple sources of knowledge jointly in real-world applications, the hybrid form of question answering over texts and tables~(TextTableQA) has been proposed and attracted more and more attention~\citep{chen2020hybridqa, chen2020open, zhu2021tat, chen2021finqa, zhao2022multihiertt, wang2022survey}. Fact reasoning~\citep{chen2020open, chen2020hybridqa} is a critical question type of TextTableQA. It requires jointly using multiple evidence from tables and texts to reasoning the answers with different operations, such as correlation (e.g., multi-hop) and aggregation (e.g., comparison). Hyperlinks among some table cells and linked passages are essential resources to establish their relationship and support the retrieval and reasoning for multi-hop questions. As shown in Figure~\ref{fig1}, answering a complex question Q1 requires jointly reasoning from textual evidence~(P1) to table evidence~([R2, Place]) and then to other table evidence~([R2, Athlete]).
    
\begin{figure*}[h]
    \centering
	\includegraphics[width=\textwidth]{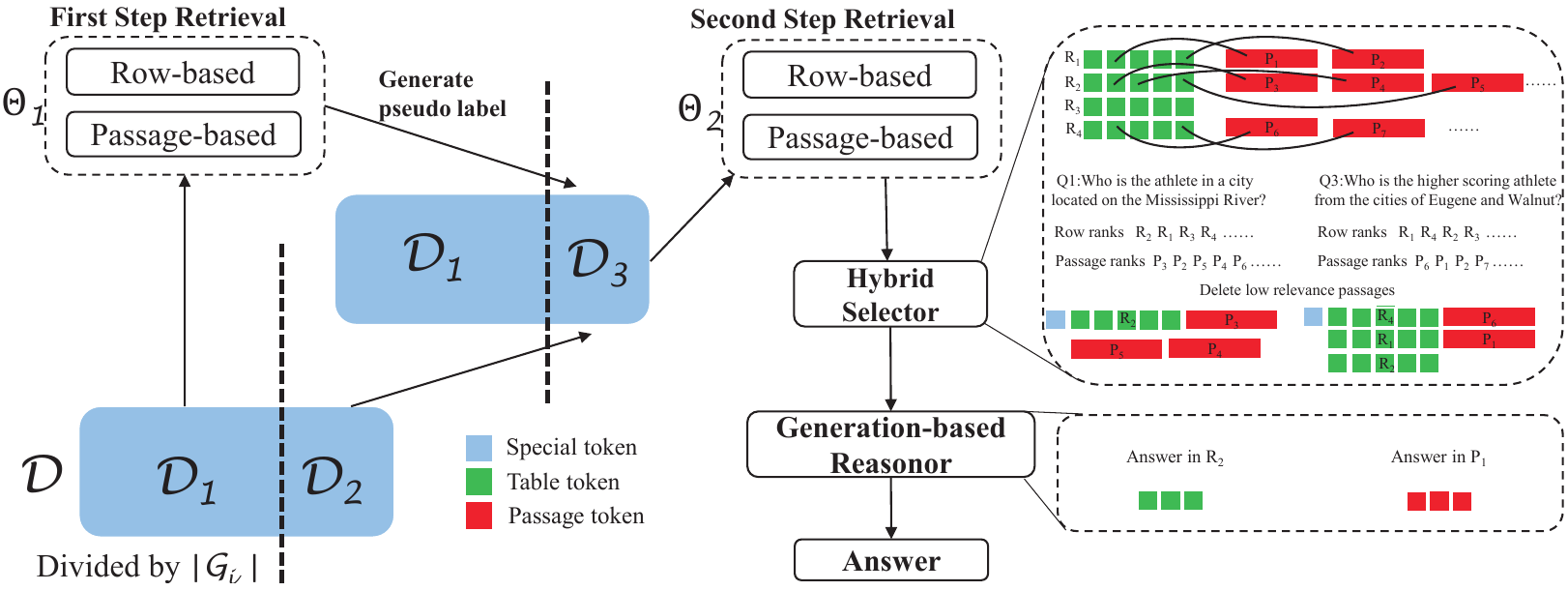}
	\caption{An overview of S$^3$HQA framework. The retrieval stage is divided into two steps. The hybrid selector considers the linked relationships between heterogeneous data to select the most relevant factual knowledge.}
	\label{fig2}
\end{figure*}  
Existing methods consist of two main stages: \emph{retriever} and \emph{reader}~\citep{chen2020hybridqa,feng2022multi}. The \emph{retriever} filters out the cells and passages with high relevance to the question, and then the \emph{reader} extracts a span from the retrieval results as the final answer. However, current methods with two stages still have three limitations as follows.

1)~\textbf{Noisy labeling for training retriever.} Existing retrieval methods usually ignore the weakly supervised answer annotation~\citep{chen2020hybridqa, wang-etal-2022-muger2, feng2022multi}. For the Q2 of Figure~\ref{fig1}, we cannot know the specific location of the hybrid evidence, only given the final answer "1960". Therefore, there is a lot of pseudo-true evidence labeled~(Marked in green) automatically by string matching, which introduces a lot of evidence noise.

2)~\textbf{Insufficient utilization of heterogeneous information.} After retrieval, existing methods selected a particular cell or passage for reading to extract the final answer~\citep{chen2020hybridqa,wang-etal-2022-muger2}. As for Q1 in Figure~\ref{fig1}, previous models were more likely to choose P1 or the coordinates [R2,Place] to extract the answer. However, these methods seldomly used the hybrid information of table schema and cell-passage hyperlinks, which is the key factor in answering multi-hop questions.
  
3)~\textbf{Deficient ability for different reasoning operations.} Previous methods~\citep{eisenschlos2021mate,kumar2021multi,wang-etal-2022-muger2} mainly used an extraction module to obtain answers, which cannot support knowledge reasoning that requires comparison, calculation, and other operations.

In this paper, we propose a three-stage approach S$^3$HQA to solve the above problems. (1)~\textbf{Retriever with Refinement Training}, we propose a two-step training method, splitting the training data into two parts, so that the noise in the retrieval phase can be alleviated. (2)~\textbf{Hybrid Selector} has been proposed and selects supporting facts with different granularity and resources depending on the question type. By considering the hybrid data of tables and text, this paper proposes a hybrid selection algorithm that can effectively utilize the heterogeneous information of tables and passages. (3)~\textbf{Generation-based reasoner} utilizes a generation-based model for addressing different question types. The model allows better aggregation of information on the input side, which not only have better multi-hop reasoning capabilities but also be able to handle comparison and counting questions. Furthermore, we are the first to use the LLM in-context learning approach for table-text hybrid question-answering tasks.

We evaluate our proposed model on the challenging TextTableQA benchmark HybridQA. The empirical results show that our approach outperforms all the existing models\footnote{We released the source code at~\url{https://github.com/lfy79001/S3HQA}}.

\section{Our Approach}

\subsection{Problem Definition}
Given a natural language question $\mathcal{Q} =\left \{ q_{i}  \right \} _{i=1}^{|\mathcal{Q}|} $ and a table $\mathcal{T}$ with $\left \langle \mathcal{H} , \mathcal{R}  \right \rangle $, $\mathcal{H}$ indicates the table headers, and $\mathcal{R}=\left \{ r_{i}  \right \} _{i=1}^{|\mathcal{R}|}$ indicates the rows with number $|\mathcal{R}|$. Each row $r_{i}$ is consists of $N$ cells $r_{i}=\left \{ c_{ij}  \right \}_{j=1}^{N}$. The header's  number is also $N$. Some cells have a linked passage $\mathcal{P}_{ij}$. Our goal aims to generate the answer $\mathcal{A}$ with model $\Theta$, which is a span from table cells or linked passage or a derivation result of counting questions.

\subsection{Retriever with Refinement Training}
The retriever aims to perform initial filtering of heterogeneous resources. However, accurately labeling the location of answers consumes high labeling costs. For TextTableQA data, the answer $\mathcal{A}$ usually appears in multiple locations, which makes it difficult for us to generate precise retrieval labels. We use a two-step training method, with a row-based retriever and a passage-based retriever for each step.

Inspired by \cite{kumar2021multi}, the retrieval has two steps. First, we divide the data $\mathcal{D}$ into two folds according to the string matching labels $G_{i}$. Specifically, for a question-answer instance, the answer $\mathcal{A}$ appears one time as $\mathcal{D}_{1}$, and the instance whose answer $\mathcal{A}$ appears multiple times as $\mathcal{D}_{2}$. Take the example in Figure~\ref{fig1}, Q1, Q3 belongs to $\mathcal{D}_{1}$ while Q2 belongs to $\mathcal{D}_{2}$. The data is organized in the form of $\mathrm{[CLS]}q_{1}q_{2}...q_{|Q|}\mathrm{[SEP]c_{i1}c_{i2}...c_{iN}[SEP]}$ or $\mathrm{[CLS]}q_{1}q_{2}...q_{|Q|}\mathrm{[SEP]p_{ij}[SEP]}$. 

In the first step, we only use $\mathcal{D}_{1}$ to train a model $\Theta_{1}$, which data are noiseless. Then in the second step, we use the trained weight $\Theta_{1}$ to train the model $\Theta_{2}$. For the input $x$, the loss function is:
$$L(\Theta_{2} ,x,\mathcal{R})=\sum_{z\in \mathcal{R}}^{} -q(z)\log_{}{p_{\Theta_{1} }(z|x) } $$
where $q(z)=p_{\Theta_{1} }(z|x,z\in \mathcal{R})$ is the probability distribution given by the model restricted to candidate rows $\mathcal{R}$ containing the answer span, taken here as a constant with zero gradients~\citep{eisenschlos2021mate}.

Meanwhile, we use a passage-based retriever to enhance the performance of a row-based retriever~(PassageFilter). Specifically, we use the passage-based retriever to obtain a prediction score of passage relevance. Based on this score, we reorder the input of the row-based retriever. It avoids the limitation on input sequence length imposed by the pre-trained model.

\subsection{Hybrid Selector}
This module needs to combine the results of the two granularity retrievers. As for this task, we consider the question type and the relationships between the table and linked passages essential. As shown in Figure~\ref{fig2}, the hybrid selector chooses the appropriate data source from the two retrieval results depending on question types. 

Specifically, for general \emph{bridge} multi-hop questions, we use a single row and its linked passage. While for \emph{comparison/count} questions, we consider multiple rows and further filter the related sentences, delete the linked paragraphs with the low scores. This not only enables the generation module to obtain accurate information, but also prevents the introduction of a large amount of unrelated information. The selector algorithm outputs a mixed sequence with high relevance based on the relationship between the question, the table, and the passages. The algorithm is shown in Algorithm~\ref{alg1}. 
\begin{algorithm}[htb]\small
        \renewcommand{\algorithmicrequire}{\textbf{Input:}}
	\renewcommand{\algorithmicensure}{\textbf{Output:}}
	\caption{Hybrid Selector Algorithm.} 
	\label{alg1} 
	\begin{algorithmic}[1]
		\REQUIRE question $\mathcal{Q}$, table rows $\mathcal{R}$, linked passages $\mathcal{P}$, row-based retriever $\Theta_{R}$, passage-based retriever $\Theta_{P}$, selector target row count $N_{S}$
		\ENSURE generator input $\mathcal{S}$\\
            \emph{Get the row/passage ordered list by relevant scores}
            \STATE $ \mathcal{O_{R}}  \gets sort(\Theta_{R}(\mathcal{Q},\mathcal{R}))  $ 
            \STATE $ \mathcal{O_{P}}  \gets sort(\Theta_{P}(\mathcal{Q},\mathcal{P}))  $
            \STATE  $\mathrm{p}^{\mathrm{type}} \gets Classification(Q) $
		\IF{$ \mathrm{p}^{\mathrm{type}} =  bridge$} 
            \IF{$ \mathcal{O_{P}}[0] $ in $\mathcal{O_{R}}[0]  $}
                \STATE $\mathcal{S} \gets \mathcal{Q} + \mathcal{O_{R}}[0]$
            \ELSE 
                \STATE $\mathcal{S} \gets \mathcal{Q} + \mathcal{O_{R}}[0] +\mathcal{O_{P}}[0] $
            \ENDIF 
		\ELSE  
                \STATE $ \mathcal{O_{PC}}  \gets \mathcal{P}[len(\mathcal{O_{P}})//2:] $
                \STATE $ \mathcal{S}  \gets \mathcal{Q} + \mathcal{O_{R}}[0:N_{S}] - \mathcal{O_{PC}} $
		\ENDIF 
            \RETURN $\mathcal{S}$
	\end{algorithmic} 
\end{algorithm}

\subsection{Generation-based Reasoner}
The results of the selector take into account both two granularity. Unlike the previous approaches, which were based on a span extraction module, we use a generation-based model for answer prediction. 

\subsubsection{Row-wise generator}
To generate an accurate answer string $\mathcal{A}=(a_{1},a_{2},...,a_{n})$ given the question $\mathcal{Q}$ and selection evidence $\mathcal{S}$, we perform lexical analysis to identify the question type, such as counting or comparison, by looking for certain keywords or comparative adjectives. We utilize two special tags $\mathrm{\left \langle Count\right \rangle}$  and $\mathrm{\left \langle Compare \right \rangle}$, which indicates the question types.

We then use the results of the passage retriever to rank the passages in order of their relevance, eliminating the impact of model input length limitations. Finally, we train a Seq2Seq language model with parameters $\Theta$, using the input sequence $\mathcal{Q}, \mathcal{S}$ and the previous outputs $a_{<i}$ to optimize the product of the probabilities of the output sequence $a_{1},a_{2},...,a_{n}$:
$$\mathcal{A}=argmax\prod_{i=1}^{n} P(a_{i}|a_{<i},\mathcal{Q}, \mathcal{S};\Theta )$$

\begin{table*}[h]
\small
\centering
\begin{tabular}{l|p{0.5cm}p{0.5cm}p{0.5cm}p{0.5cm}|p{0.5cm}p{0.5cm}p{0.5cm}p{0.5cm}|p{0.5cm}p{0.5cm}p{0.5cm}p{0.5cm}}
\hline
                  & \multicolumn{4}{c|}{\textbf{Table}}                                   & \multicolumn{4}{c|}{\textbf{Passage}}                                 & \multicolumn{4}{c}{\textbf{Total}}                                   \\
                  & \multicolumn{2}{c}{\textbf{Dev}} & \multicolumn{2}{c|}{\textbf{Test}} & \multicolumn{2}{c}{\textbf{Dev}} & \multicolumn{2}{c|}{\textbf{Test}} & \multicolumn{2}{c}{\textbf{Dev}} & \multicolumn{2}{c}{\textbf{Test}} \\
                  & EM              & F1             & EM               & F1              & EM              & F1             & EM               & F1              & EM              & F1             & EM              & F1              \\ \hline
Unsupervised-QG~\citep{pan2021unsupervised}   & -               & -              & -                & -               & -               & -              & -                & -               & 25.7            & 30.5           & -               & -               \\
HYBRIDER~\citep{chen2020hybridqa}          & 54.3            & 61.4           & 56.2             & 63.3            & 39.1            & 45.7           & 37.5             & 44.4            & 44.0            & 50.7           & 43.8            & 50.6            \\
DocHopper~\citep{sun2021end}         & -               & -              & -                & -               & -               & -              & -                & -               & 47.7            & 55.0           & 46.3            & 53.3            \\
MuGER$^2$~\citep{wang-etal-2022-muger2}         & 60.9            & 69.2           & 58.7             & 66.6            & 56.9            & 68.9           & 57.1             & 68.6            & 57.1            & 67.3           & 56.3            & 66.2            \\
POINTR~\citep{eisenschlos2021mate}           & 68.6            & 74.2           & 66.9             & 72.3            & 62.8            & 71.9           & 62.8             & 71.9            & 63.4            & 71.0           & 62.8            & 70.2            \\
DEHG~\citep{feng2022multi}              & -               & -              & -                & -               & -               & -              & -                & -               & 65.2            & \textbf{76.3}  & 63.9            & 75.5            \\
MITQA~\citep{kumar2021multi}             & 68.1            & 73.3           & 68.5             & 74.4            & 66.7            & 75.6           & 64.3             & 73.3            & 65.5            & 72.7           & 64.3            & 71.9            \\
MAFiD~\citep{lee2023mafid}             & 69.4            & 75.2           & 68.5             & 74.9            & 66.5            & 75.5           & 65.7             & 75.3            & 66.2            & 74.1           & 65.4            & 73.6            \\ \hline \hline
\textbf{S$^3$HQA} & \textbf{70.3}   & \textbf{75.3}  & \textbf{70.6}    & \textbf{76.3}   & \textbf{69.9}   & \textbf{78.2}  & \textbf{68.7}    & \textbf{77.8}   & \textbf{68.4}   & 75.3           & \textbf{67.9}   & \textbf{75.5}   \\ \hline
Human             & -               & -              & -                & -               & -               & -              & -                & -               & -               & -              & 88.2             & 93.5               \\ \hline
\end{tabular}
\caption{Performance of our model and related work on the HybridQA dataset.
}
\label{tab:main_result}
\end{table*}

\subsubsection{LLM prompting generator}
With the emergence of large language models, In-Context Learning \cite{dong2022survey} and Chain-of-Thought prompting \cite{wei2022chain} have become two particularly popular research topics in this field. In this paper, we introduce a prompting strategy for multi-hop TextTableQA.

We utilize selection evidence $\mathcal{S}$ and apply LLM-based prompting. We conducted experiments on both vanilla prompting and chain-of-thought prompting in zero-shot and few-shot scenarios.


\section{Experiment}
\subsection{Experiment Setup}
\textbf{Datasets} We conduct experiments on HybridQA~\citep{chen2020hybridqa}. The detailed statistics are shown in Appendix~\ref{sec:appendix}. For evaluation, we followed the official evaluation to report exact match accuracy and F1 score. \\
\textbf{Implementation details} The implementation details are shown in Appendix~\ref{imple:appendix}. The experimental results are the average of five times results.

\subsection{Fully-supervised Results}
Table~\ref{tab:main_result} shows the comparison results between our models with previous typical approaches on both development and test sets. It shows that our proposed S$^3$HQA works significantly better than the baselines in terms of EM and F1 on HybridQA. The results indicate that S$^3$HQA is an effective model for multi-hop question answering over tabular and textual data. Specifically, it can effectively handle multi-hop reasoning and make full use of heterogeneous information.

However, we found that our approach was outperformed by the DEHG model~\citep{feng2022multi} in terms of F1 score on the Dev set. We speculate that this might be because the DEHG approach uses their own Open Information Extraction (OIE) tool.


\begin{table}[h]
\centering
\begin{tabular}{lll}
\hline
\multicolumn{1}{c}{\multirow{2}{*}{Model}} & \multicolumn{2}{c}{Dev}                             \\
\multicolumn{1}{c}{}                       & \multicolumn{1}{c}{EM}   & \multicolumn{1}{c}{F1}   \\ \hline
\multicolumn{3}{c}{Zero-shot prompt}                                                             \\ \hline
GPT3.5 direct                              & \multicolumn{1}{c}{33.1} & \multicolumn{1}{c}{50.5} \\
GPT3.5 CoT                                 & 52.9                     & 66.6                     \\ \hline
\multicolumn{3}{c}{Few-shot prompt~(2-shot)}                                                \\ \hline
GPT3.5 direct                              & 57.1                     & 68.8                     \\
GPT3.5 CoT                                 & \textbf{60.3}         & \textbf{72.1}                     \\ \hline
\end{tabular}
\caption{Performance Comparison of LLM-Prompting Method on Zero-Shot and Few-Shot Scenarios for HybridQA Dataset.}
\label{tab:llm_prompt}
\end{table}

\subsection{LLM-prompting Results}
\label{llm_prompt}
We present our zero-shot and few-shot results in Table~\ref{tab:llm_prompt}. "\textbf{Direct}" refers to a simple prompting method where only the question, context, and answer are provided to the model without any additional reasoning process. In contrast, "\textbf{CoT}" involves a human-authored Chain-of-Thought reasoning process that provides a more structured and logical way of prompting the model. The experiments demonstrate that in-context learning used to prompt large language models can achieve promising results. Specifically, utilizing the Chain-of-Thought prompt method can significantly enhance the model's performance. 

However, it's worth noting that there is still a performance gap compared to fine-tuning the model on the full dataset~(Table~\ref{tab:main_result}). Fine-tuning allows the model to learn more specific information about the TextTableQA task, resulting in better performance. Nevertheless, our results show that the LLM-prompting method can be a useful alternative to fine-tuning, especially when there is a limited amount of labeled data available.

\subsection{Ablation Studies}
We conduct ablation studies on the test set. We validate the effects of three modules: \emph{retriever with refinement training}, \emph{hybrid selector}, and \emph{generation-based reasoner}. The retriever performs initial filtering of heterogeneous resources; Selectors combined with hyperlinks further identify the exact evidence needed to answer multi-hop questions; and the reasoner uses the selection evidence to obtain the final answer.\\

\begin{table}[h]
\centering
\begin{tabular}{ll}
\hline
Model               & Top1  \\ \hline
$\textrm{S}^{3}\textrm{HQA-Retriever}_{\textrm{DB}}$      & \textbf{88.0}    \\
$\textrm{S}^{3}\textrm{HQA-Retriever}_{\textrm{BE}}$       & 87.3    \\
w/o Refinement training  & 84.1                 \\
w/o PassageFilter  & 85.3                  \\
$\textrm{Vanilla-Retriever}_{\textrm{BE}}$         & 82.0            \\
\hline
\end{tabular}
\caption{Ablation study of retrieval results. DB and BE denote models based on Deberta-base~\citep{he2020deberta} and BERT-base-uncased~\citep{devlin2018bert}, respectively}
\label{tab:ablation_study1}
\end{table}

\begin{table}[h]
\centering
\begin{tabular}{lll}
\hline
Model               & EM          & F1 \\ \hline
S$^3$HQA                 & \textbf{67.9}  & \textbf{76.5}         \\
w/o hybrid selector        & 65.0        & 74.9          \\
w/o special tags        & 67.2        & 76.0          \\
BERT-large reader    & 66.8        & 75.8           \\ \hline
\end{tabular}
\caption{Ablation study of S$^3$HQA.}
\label{tab:ablation_study2}
\end{table}
\textbf{Effect of proposed retriever.} As shown in the Table~\ref{tab:ablation_study1}, under the setting of using the BERT-base-uncased model, sing the BERT-base-uncased model setting, the retriever with \emph{refinement training} achieved 87.2. When we use Deberta-base, the top1 retrieval performance improved by 0.8$\%$. For \emph{w/o refinement training}, we use the entire data directly for training, the top1 recall drops about 3.2$\%$. For \emph{w/o PassageFilter}, we remove the mechanism, the top1 recall drops about 3.2$\%$. For \emph{Vanilla-Retriever}, we use the row-based retriever~\citep{kumar2021multi} and remove all our mechanisms, the top1 score drops about 5.3$\%$. This shows that our model can solve the weakly supervised data noise problem well.

\textbf{Effect of hybrid selector.} As shown in the Table~\ref{tab:ablation_study2}, we removed the selector of S$^3$HQA and replaced it with the previous cell-based selector~\citep{wang-etal-2022-muger2}. This method directly uses the top1 result of the row retriever as input to the generator. \emph{w/o hybrid selector} shows that the EM drops 2.9$\%$ and F1 drops 1.6$\%$, which proves the effectiveness of our selector approach.

\textbf{Effect of reasoner.} As shown in the Table~\ref{tab:ablation_study2}, we design two baselines. \emph{BERT-large reader}~\citep{chen2020hybridqa,wang-etal-2022-muger2} uses BERT~\citep{devlin2018bert} as encoder and solves this task by predicting the start/end tokens. \emph{w/o special tags} deletes the special tags. Both the two experiments demonstrate our S$^3$HQA reasoner performs the best for HybridQA task. \\

\section{Related Work}
The TextTableQA task~\citep{wang2022survey} has attracted more and more attention. As for multi-hop type dataset, previous work used pipeline approach~\citep{chen2020hybridqa}, unsupervised approach~\citep{pan2021unsupervised}, multi-granularity~\citep{wang-etal-2022-muger2}, table pre-trained language model~\citep{eisenschlos2021mate}, multi-instance learning~\citep{kumar2021multi} and graph neural network~\citep{feng2022multi} to solve this task. As for numerical reasoning task, which is quite different from multi-hop type dataset, there is also a lot of work~\citep{zhu2021tat, zhao2022multihiertt, zhou2022unirpg, lei2022answering, li2022dyrren, wei2023multi} to look at these types of questions. Unlike these methods, our proposed three-stage model S$^3$HQA can alleviate noises from weakly supervised and solve different types of multi-hop TextTableQA questions by handling the relationship between tables and text.

\section{Conclusion}
This paper proposes a three-stage model consisting of retriever, selector, and reasoner, which can effectively address multi-hop TextTableQA. The proposed method solves three drawbacks of the previous methods: noisy labeling for training retriever, insufficient utilization of heterogeneous information, and deficient ability for reasoning. It achieves new state-of-the-art performance on the widely used benchmark HybridQA. In future work, we will design more interpretable TextTableQA models to predict the explicit reasoning path.

\section*{Limitations}
Since the multi-hop TextTableQA problem has only one dataset HybridQA, our model has experimented on only one dataset. This may lead to a lack of generalizability of our model. Transparency and interpretability are important in multi-hop question answering. While our model achieves the best results, the model does not fully predict the reasoning path explicitly and can only predict the row-level path and passage-level path. In future work, we will design more interpretable TextTableQA models.

\bibliography{custom}

\begin{thebibliography}{27}
\expandafter\ifx\csname natexlab\endcsname\relax\def\natexlab#1{#1}\fi

\bibitem[{Bird(2006)}]{bird2006nltk}
Steven Bird. 2006.
\newblock Nltk: the natural language toolkit.
\newblock In \emph{Proceedings of the COLING/ACL 2006 Interactive Presentation
  Sessions}, pages 69--72.

\bibitem[{Chen et~al.(2020{\natexlab{a}})Chen, Chang, Schlinger, Wang, and
  Cohen}]{chen2020open}
Wenhu Chen, Ming-Wei Chang, Eva Schlinger, William~Yang Wang, and William~W
  Cohen. 2020{\natexlab{a}}.
\newblock Open question answering over tables and text.
\newblock In \emph{International Conference on Learning Representations}.

\bibitem[{Chen et~al.(2020{\natexlab{b}})Chen, Zha, Chen, Xiong, Wang, and
  Wang}]{chen2020hybridqa}
Wenhu Chen, Hanwen Zha, Zhiyu Chen, Wenhan Xiong, Hong Wang, and William~Yang
  Wang. 2020{\natexlab{b}}.
\newblock Hybridqa: A dataset of multi-hop question answering over tabular and
  textual data.
\newblock In \emph{Findings of the Association for Computational Linguistics:
  EMNLP 2020}, pages 1026--1036.

\bibitem[{Chen et~al.(2021)Chen, Chen, Smiley, Shah, Borova, Langdon, Moussa,
  Beane, Huang, Routledge et~al.}]{chen2021finqa}
Zhiyu Chen, Wenhu Chen, Charese Smiley, Sameena Shah, Iana Borova, Dylan
  Langdon, Reema Moussa, Matt Beane, Ting-Hao Huang, Bryan~R Routledge, et~al.
  2021.
\newblock Finqa: A dataset of numerical reasoning over financial data.
\newblock In \emph{Proceedings of the 2021 Conference on Empirical Methods in
  Natural Language Processing}, pages 3697--3711.

\bibitem[{Devlin et~al.(2018)Devlin, Chang, Lee, and
  Toutanova}]{devlin2018bert}
Jacob Devlin, Ming-Wei Chang, Kenton Lee, and Kristina Toutanova. 2018.
\newblock Bert: Pre-training of deep bidirectional transformers for language
  understanding.
\newblock \emph{arXiv preprint arXiv:1810.04805}.

\bibitem[{Dong et~al.(2022)Dong, Li, Dai, Zheng, Wu, Chang, Sun, Xu, and
  Sui}]{dong2022survey}
Qingxiu Dong, Lei Li, Damai Dai, Ce~Zheng, Zhiyong Wu, Baobao Chang, Xu~Sun,
  Jingjing Xu, and Zhifang Sui. 2022.
\newblock A survey for in-context learning.
\newblock \emph{arXiv preprint arXiv:2301.00234}.

\bibitem[{Eisenschlos et~al.(2021)Eisenschlos, Gor, Mueller, and
  Cohen}]{eisenschlos2021mate}
Julian Eisenschlos, Maharshi Gor, Thomas Mueller, and William Cohen. 2021.
\newblock Mate: Multi-view attention for table transformer efficiency.
\newblock In \emph{Proceedings of the 2021 Conference on Empirical Methods in
  Natural Language Processing}, pages 7606--7619.

\bibitem[{Feng et~al.(2022)Feng, Han, Sun, and Li}]{feng2022multi}
Yue Feng, Zhen Han, Mingming Sun, and Ping Li. 2022.
\newblock Multi-hop open-domain question answering over structured and
  unstructured knowledge.
\newblock In \emph{Findings of the Association for Computational Linguistics:
  NAACL 2022}, pages 151--156.

\bibitem[{He et~al.(2020)He, Liu, Gao, and Chen}]{he2020deberta}
Pengcheng He, Xiaodong Liu, Jianfeng Gao, and Weizhu Chen. 2020.
\newblock Deberta: Decoding-enhanced bert with disentangled attention.
\newblock \emph{arXiv preprint arXiv:2006.03654}.

\bibitem[{Kumar et~al.(2021)Kumar, Chemmengath, Gupta, Sen, Bharadwaj, and
  Chakrabarti}]{kumar2021multi}
Vishwajeet Kumar, Saneem Chemmengath, Yash Gupta, Jaydeep Sen, Samarth
  Bharadwaj, and Soumen Chakrabarti. 2021.
\newblock Multi-instance training for question answering across table and
  linked text.
\newblock \emph{arXiv preprint arXiv:2112.07337}.

\bibitem[{Lee et~al.(2023)Lee, Park, Seo, Jeon, Kang, and Na}]{lee2023mafid}
Sung-Min Lee, Eunhwan Park, Daeryong Seo, Donghyeon Jeon, Inho Kang, and
  Seung-Hoon Na. 2023.
\newblock Mafid: Moving average equipped fusion-in-decoder for question
  answering over tabular and textual data.
\newblock In \emph{Findings of the Association for Computational Linguistics:
  EACL 2023}, pages 2292--2299.

\bibitem[{Lei et~al.(2022)Lei, He, Li, Zhao, and Liu}]{lei2022answering}
Fangyu Lei, Shizhu He, Xiang Li, Jun Zhao, and Kang Liu. 2022.
\newblock Answering numerical reasoning questions in table-text hybrid contents
  with graph-based encoder and tree-based decoder.
\newblock In \emph{Proceedings of the 29th International Conference on
  Computational Linguistics}, pages 1379--1390.

\bibitem[{Lewis et~al.(2020)Lewis, Liu, Goyal, Ghazvininejad, Mohamed, Levy,
  Stoyanov, and Zettlemoyer}]{lewis2020bart}
Mike Lewis, Yinhan Liu, Naman Goyal, Marjan Ghazvininejad, Abdelrahman Mohamed,
  Omer Levy, Veselin Stoyanov, and Luke Zettlemoyer. 2020.
\newblock Bart: Denoising sequence-to-sequence pre-training for natural
  language generation, translation, and comprehension.
\newblock In \emph{Proceedings of the 58th Annual Meeting of the Association
  for Computational Linguistics}, pages 7871--7880.

\bibitem[{Li et~al.(2022)Li, Zhu, Liu, Ju, Qu, and Cheng}]{li2022dyrren}
Xiao Li, Yin Zhu, Sichen Liu, Jiangzhou Ju, Yuzhong Qu, and Gong Cheng. 2022.
\newblock Dyrren: A dynamic retriever-reranker-generator model for numerical
  reasoning over tabular and textual data.
\newblock \emph{arXiv preprint arXiv:2211.12668}.

\bibitem[{Pan et~al.(2021)Pan, Chen, Xiong, Kan, and
  Wang}]{pan2021unsupervised}
Liangming Pan, Wenhu Chen, Wenhan Xiong, Min-Yen Kan, and William~Yang Wang.
  2021.
\newblock Unsupervised multi-hop question answering by question generation.
\newblock In \emph{Proceedings of the 2021 Conference of the North American
  Chapter of the Association for Computational Linguistics: Human Language
  Technologies}, pages 5866--5880.

\bibitem[{Pasupat and Liang(2015)}]{pasupat2015compositional}
Panupong Pasupat and Percy Liang. 2015.
\newblock Compositional semantic parsing on semi-structured tables.
\newblock In \emph{Proceedings of the 53rd Annual Meeting of the Association
  for Computational Linguistics and the 7th International Joint Conference on
  Natural Language Processing (Volume 1: Long Papers)}, pages 1470--1480.

\bibitem[{Paszke et~al.(2019)Paszke, Gross, Massa, Lerer, Bradbury, Chanan,
  Killeen, Lin, Gimelshein, Antiga et~al.}]{paszke2019pytorch}
Adam Paszke, Sam Gross, Francisco Massa, Adam Lerer, James Bradbury, Gregory
  Chanan, Trevor Killeen, Zeming Lin, Natalia Gimelshein, Luca Antiga, et~al.
  2019.
\newblock Pytorch: An imperative style, high-performance deep learning library.
\newblock \emph{Advances in neural information processing systems}, 32.

\bibitem[{Rajpurkar et~al.(2016)Rajpurkar, Zhang, Lopyrev, and
  Liang}]{rajpurkar2016squad}
Pranav Rajpurkar, Jian Zhang, Konstantin Lopyrev, and Percy Liang. 2016.
\newblock Squad: 100,000+ questions for machine comprehension of text.
\newblock In \emph{Proceedings of the 2016 Conference on Empirical Methods in
  Natural Language Processing}, pages 2383--2392.

\bibitem[{Sun et~al.(2021)Sun, Cohen, and Salakhutdinov}]{sun2021end}
Haitian Sun, William~W Cohen, and Ruslan Salakhutdinov. 2021.
\newblock End-to-end multihop retrieval for compositional question answering
  over long documents.

\bibitem[{Wang et~al.(2022{\natexlab{a}})Wang, Dou, and Che}]{wang2022survey}
Dingzirui Wang, Longxu Dou, and Wanxiang Che. 2022{\natexlab{a}}.
\newblock A survey on table-and-text hybridqa: Concepts, methods, challenges
  and future directions.
\newblock \emph{arXiv preprint arXiv:2212.13465}.

\bibitem[{Wang et~al.(2022{\natexlab{b}})Wang, Bao, Duan, Wu, He, and
  Zhao}]{wang-etal-2022-muger2}
Yingyao Wang, Junwei Bao, Chaoqun Duan, Youzheng Wu, Xiaodong He, and Tiejun
  Zhao. 2022{\natexlab{b}}.
\newblock \href {https://aclanthology.org/2022.findings-emnlp.498} {{M}u{GER}2:
  Multi-granularity evidence retrieval and reasoning for hybrid question
  answering}.
\newblock In \emph{Findings of the Association for Computational Linguistics:
  EMNLP 2022}, pages 6687--6697, Abu Dhabi, United Arab Emirates. Association
  for Computational Linguistics.

\bibitem[{Wei et~al.(2022)Wei, Wang, Schuurmans, Bosma, Chi, Le, and
  Zhou}]{wei2022chain}
Jason Wei, Xuezhi Wang, Dale Schuurmans, Maarten Bosma, Ed~Chi, Quoc Le, and
  Denny Zhou. 2022.
\newblock Chain of thought prompting elicits reasoning in large language
  models.
\newblock \emph{arXiv preprint arXiv:2201.11903}.

\bibitem[{Wei et~al.(2023)Wei, Lei, Zhang, Zhao, and Liu}]{wei2023multi}
Yifan Wei, Fangyu Lei, Yuanzhe Zhang, Jun Zhao, and Kang Liu. 2023.
\newblock Multi-view graph representation learning for answering hybrid
  numerical reasoning question.
\newblock \emph{arXiv preprint arXiv:2305.03458}.

\bibitem[{Yu et~al.(2018)Yu, Zhang, Yang, Yasunaga, Wang, Li, Ma, Li, Yao,
  Roman et~al.}]{yu2018spider}
Tao Yu, Rui Zhang, Kai Yang, Michihiro Yasunaga, Dongxu Wang, Zifan Li, James
  Ma, Irene Li, Qingning Yao, Shanelle Roman, et~al. 2018.
\newblock Spider: A large-scale human-labeled dataset for complex and
  cross-domain semantic parsing and text-to-sql task.
\newblock In \emph{Proceedings of the 2018 Conference on Empirical Methods in
  Natural Language Processing}, pages 3911--3921.

\bibitem[{Zhao et~al.(2022)Zhao, Li, Li, and Zhang}]{zhao2022multihiertt}
Yilun Zhao, Yunxiang Li, Chenying Li, and Rui Zhang. 2022.
\newblock Multihiertt: Numerical reasoning over multi hierarchical tabular and
  textual data.
\newblock In \emph{Proceedings of the 60th Annual Meeting of the Association
  for Computational Linguistics (Volume 1: Long Papers)}, pages 6588--6600.

\bibitem[{Zhou et~al.(2022)Zhou, Bao, Duan, Wu, He, and Zhao}]{zhou2022unirpg}
Yongwei Zhou, Junwei Bao, Chaoqun Duan, Youzheng Wu, Xiaodong He, and Tiejun
  Zhao. 2022.
\newblock Unirpg: Unified discrete reasoning over table and text as program
  generation.
\newblock \emph{arXiv preprint arXiv:2210.08249}.

\bibitem[{Zhu et~al.(2021)Zhu, Lei, Huang, Wang, Zhang, Lv, Feng, and
  Chua}]{zhu2021tat}
Fengbin Zhu, Wenqiang Lei, Youcheng Huang, Chao Wang, Shuo Zhang, Jiancheng Lv,
  Fuli Feng, and Tat-Seng Chua. 2021.
\newblock Tat-qa: A question answering benchmark on a hybrid of tabular and
  textual content in finance.
\newblock In \emph{Proceedings of the 59th Annual Meeting of the Association
  for Computational Linguistics and the 11th International Joint Conference on
  Natural Language Processing (Volume 1: Long Papers)}, pages 3277--3287.

\end{thebibliography}
\bibliographystyle{acl_natbib}

\appendix

\section{HybridQA Dataset}
\label{sec:appendix}
HybridQA is a large-scale, complex, and multi-hop TextTableQA benchmark. Tables and texts are crawled from Wikipedia. Each row in the table describes several attributes of an instance. Each table has its hyperlinked Wikipedia passages that describe the detail of attributes.  It contains 62,682 instances in the train set, 3466 instances in the dev set and 3463 instances in the test set.

\begin{table}[!ht]
\small
\centering
\begin{tabular}{lccccc}
\toprule
Split & Train & Dev & Test & Total \\
\midrule
In-Passage  &  35,215 & 2,025  & 20,45  & 39,285 (56.4\%) \\
In-Table & 26,803 & 1,349 & 1,346 & 29,498 (42.3\%)  \\
Computed & 664  & 92 & 72 & 828 (1.1\%) \\
Total & 62,682 & 3,466 & 3,463 & 69,611 \\
\bottomrule
\end{tabular}
\caption{Data Split: In-Table means the answer comes from plain text in the table, and In-Passage means the answer comes from certain passage. }
\label{tab:data_split}
\vspace{-2ex}
\end{table}

\section{Implementation Details}
\label{imple:appendix}
\subsection{Fully-supervised Setting}
We utilize PyTorch~\citep{paszke2019pytorch} to implement our proposed model. During pre-processing, the input of questions, tables and passages are tokenized and lemmatized with the NLTK~\citep{bird2006nltk} toolkit. We conducted the experiments on a single NVIDIA GeForce RTX 3090. 

In the retriever stage, we use BERT-base-uncased~\citep{devlin2018bert} and Deberta-base~\citep{he2020deberta} to obtain the initial representations. For the first step, batch size is 1, epoch number is 5, learning rate is 7e-6~(selected from 1e-5, 7e-6, 5e-6). The training process may take around 10 hours. For the second step, we use a smaller learning rate 2e-6~(selected from 5e-6, 3e-6, 2e-6), epoch number is 5. The training process may take around 8 hours. In the selector stage, target row count $N_{S}$ is 3. In the generator stage, we use BART-large language model~\citep{lewis2020bart}, the learning rate is 1e-5~(selected from 5e-5, 1e-5, 5e-6), batch size is 8, epoch number is 10, beam size is 3 and max generate length is 20. 

\subsection{LLM-prompting Setting}
We use the OpenAI GPT-3.5~(text-davinci-003) API model with the setting $temperature=0$ in our experiments. For the few-shot setting, we use 2 shots. To elicit the LLM's capability to perform multi-hop reasoning, we use the text "Read the following table and text information, answer a question. Let’s think step by step." as our prompt.

\end{document}